\documentclass[10pt,twocolumn,letterpaper]{article}

\usepackage{iccv}
\usepackage{times}
\usepackage{epsfig}
\usepackage{graphicx}
\usepackage{amsmath}
\usepackage{amssymb}
\usepackage{booktabs}

\usepackage{caption}
\usepackage{subcaption}

\usepackage[breaklinks=true,bookmarks=false]{hyperref}

\iccvfinalcopy 

\def\iccvPaperID{7} 

\ificcvfinal\fi

\begin{document}

\title{VAST: Vivify Your Talking Avatar via Zero-Shot Expressive Facial Style Transfer}

\author{Liyang Chen$^{1,3}$, Zhiyong Wu$^{1}$, Runnan Li$^{3}$, Weihong Bao$^{1}$, Jun Ling$^{2}$, Xu Tan$^{3}$, Sheng Zhao$^{3}$ \\
$^1$Shenzhen International Graduate School, Tsinghua University\\
$^2$Shanghai Jiao Tong University \quad $^3$Microsoft \\
{\tt\small \{cly21, bwh21\}$@$mails.tsinghua.edu.cn zywu@sz.tsinghua.edu.cn} \\
{\tt\small lingjun$@$sjtu.edu.cn \{runnan.li, xuta, sheng.zhao\}$@$microsoft.com}}

\maketitle
\ificcvfinal\thispagestyle{empty}\fi

\begin{abstract}
Current talking face generation methods mainly focus on speech-lip synchronization. However, insufficient investigation on the facial talking style leads to a lifeless and monotonous avatar.
Most previous works fail to imitate expressive styles from arbitrary video prompts and ensure the authenticity of the generated video. 
This paper proposes an unsupervised variational style transfer model (VAST) to vivify the neutral photo-realistic avatars.
Our model consists of three key components: a style encoder that extracts facial style representations from the given video prompts; a hybrid facial expression decoder to model accurate speech-related movements; a variational style enhancer that enhances the style space to be highly expressive and meaningful. 
With our essential designs on facial style learning, our model is able to flexibly capture the expressive facial style from arbitrary video prompts and transfer it onto a personalized image renderer in a zero-shot manner. 
Experimental results demonstrate the proposed approach contributes to a more vivid talking avatar with higher authenticity and richer expressiveness.
\end{abstract}

\section{Introduction}

Building audio-driven photo-realistic avatars to provide humanoid and natural interaction experiences for users 
is highly attractive. This technology has great potential in various scenarios, such as human-computer interaction, virtual reality, filmmaking, game creation, and online education. 
While previous works \cite{obama_2017, atvg, nvp_2020,duallip_2020, wav2lip_2020, zhang2021facial, synctalkingface_2022} have achieved great strides in generating high-quality avatar videos with speech-aligned lip movements, the lack of expressive facial expressions still limits their widespread use. Facial expressions convey more information beyond the speech context and can make the speech more persuasive, encouraging, and appealing. This creates a vivid avatar other than a neutral talking avatar.

\begin{figure}[t]
\centering
\includegraphics[width=0.9\linewidth]{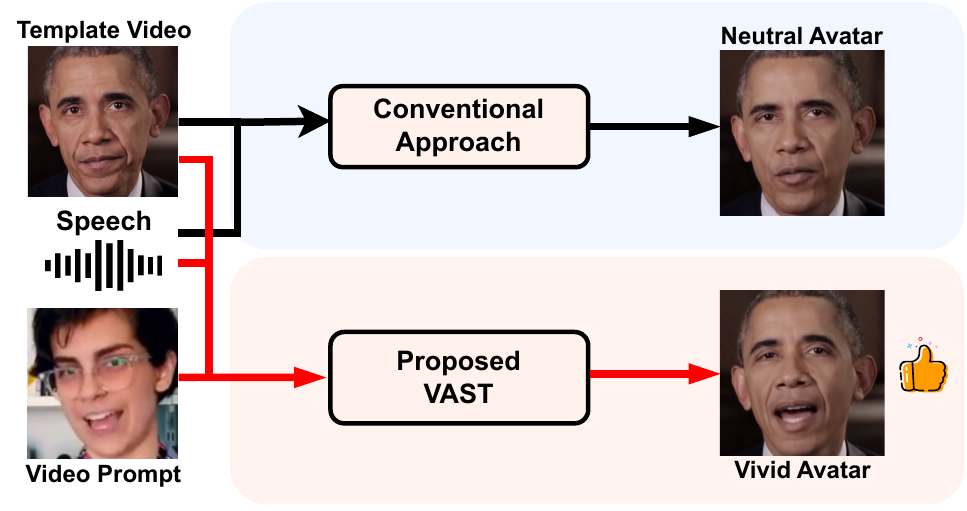}
\caption{Concept diagram for the proposed method. Expressive video prompt is employed to amend the expression prediction for vivid avatar generation.}
\label{overall}
\end{figure}

Several recent approaches \cite{mead_2020, evp_2021, wu_2021, eamm_2022} aim to generate vivid avatars by modeling facial talking style information.  For instance, \cite{mead_2020, evp_2021} introduce discrete emotional labels to provide explicit style guidance. StyleAvatar \cite{wu_2021} represents the facial styles with manually-defined features. GC-AVT \cite{gavc_2022} and EAMM \cite{eamm_2022}  disentangles style from the raw reference video images.
Although these methods can improve the expressiveness of generated avatars, they still suffer from certain limitations: 1) They struggle to transfer natural facial styles from arbitrary video in a robust manner; 2) The style representation derived by these methods cannot effectively preserve the style being imitated, resulting in the deficient expressiveness of the synthesized avatar; 3) These methods compromise the authenticity of the generation, particularly in terms of speech-lip synchronization and naturalness.

In this paper, we propose an unsupervised \textbf{VA}riational \textbf{S}tyle \textbf{T}ransfer model (VAST) to vivify the neutral photo-realistic avatar with arbitrary video prompts.
As shown in Fig.~\ref{overall}, compared to conventional approaches, the proposed model utilizes the video prompts as an additional input, alongside a speech utterance and a template video of the target avatar, to generate a vivid avatar that reenacts speech-synchronized mouth movements with facial style that transferred from the given video prompt.

The proposed VAST designs an unsupervised encoder-decoder architecture to learn facial style representation during training, and transfer the style in a zero-shot manner during inference. The encoder, based on the convolutional and recurrent network, obtains robust style representation from variable-length expression sequences.  To further enhance the expressiveness of the learned representation, a variational autoencoder-based style enhancer is incorporated, that utilizes normalizing flow to enrich the diagonal posterior.  During decoding, VAST designs a hybrid decoder constructed with the autoregressive (AR) and non-autoregressive (NAR) networks, that separately estimate the speech-weakly-related and speech-strongly-related expression parameters. Specially, to ensure flexible facial style transfer, a parametric face model \cite{moai_2021} is employed for robust facial parameters extraction. In the end, a pretrained image renderer is utilized to synthesize visual appearances for the photo-realistic avatar. 
Extensive experiments have demonstrated that VAST outperforms state-of-the-art methods with higher authenticity and richer expressiveness in vivid avatar video generation. In expressiveness user study, the proposed VAST achieves a relative improvement of 14.4$\%$ comparing to state-of-the-art approaches.

The contributions of this work can be summarized as:
\textbf{1)} We propose a variational style transfer model and effectively transfer arbitrary expressive facial style onto a neutral avatar to produce vivid results in a zero-shot manner.
\textbf{2)} The proposed variational style enhancer obtains more expressive generation. 
\textbf{3)} The proposed hybrid decoder guarantees the speech-lip synchronization and overall authenticity.

\begin{figure*}[t]
\centering
\includegraphics[width=0.95\textwidth]{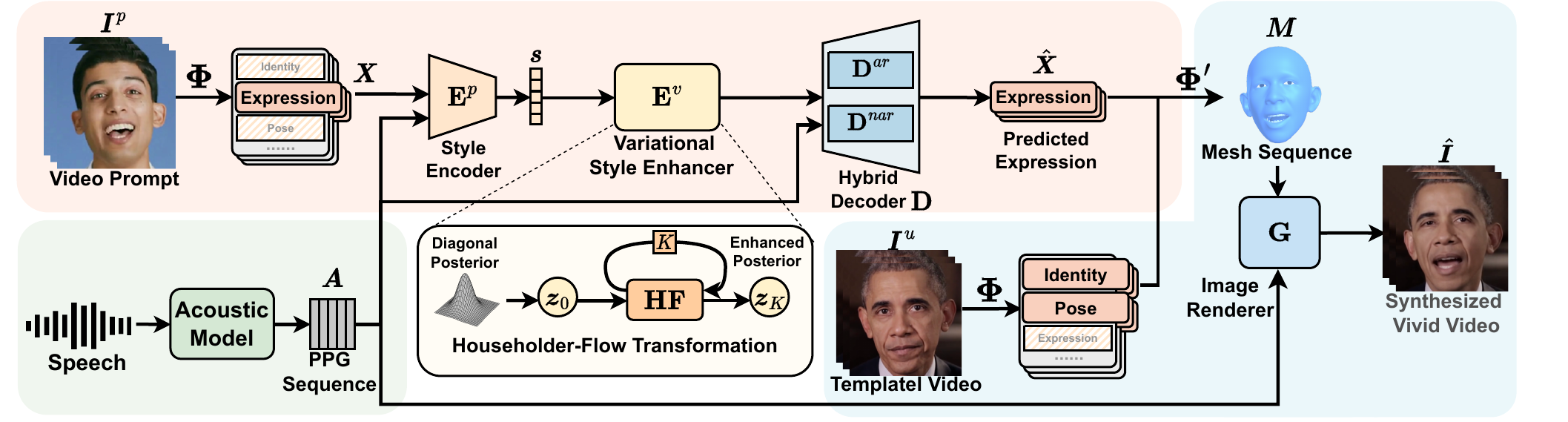} 
\caption{Overview of the proposed method. $\mathbf{\Phi}$ denotes the encoding process of the parametric face model \cite{moai_2021} that extracts the facial parameters. $\mathbf{\Phi}^{\prime}$ denotes the decoding process to output mesh from the facial parameters. In training, the video prompt and speech are paired. In inference, the video prompt can come from any source.}
\label{fig: overview}
\end{figure*}

\section{Related Works}
The audio-driven talking avatar generation task aims to generate talking videos from the given speech. We review the related works from two aspects, including the general pipelines of talking avatar and style modeling in talking avatar.

\noindent
\textbf{Audio-driven talking avatar.} 
The generation pipelines of audio-driven talking avatar can be divided into two categories: image-warping and template-editing.
The image-warping pipeline \cite{atvg, duallip_2020, wav2lip_2020, livesp_2021, livesp_2021, zhou_2021, synctalkingface_2022} aims at building a general model to drive a single portrait image. DualLip \cite{duallip_2020} and Wav2Lip \cite{wav2lip_2020} directly learn a repairment from audio for the cropped mouth region. PC-AVS \cite{zhou_2021} implicitly disentangles the audio and visual representations for pose control. SyncTalkFace \cite{synctalkingface_2022} retrieves image features with the memory-aligned audio features for finer details. LSP \cite{livesp_2021} presents a live system that generates personalized talking-head animation. AVCT \cite{avct_2022} warps the image by inferring the keypoint based dense motion fields. 
The template-editing pipeline \cite{obama_2017, obamanet_2018, nvp_2020, dvp_2020, lipsync3d_2021, stableface_2022, ye_2023} aims at building a person-specific image renderer.
Early attempts \cite{obama_2017, obamanet_2018} leverage numerous video footage of a specific person. Recent works \cite{nvp_2020, dvp_2020} introduce the 3D structural intermediate information \cite{deng_2019} into the video-based pipeline and reduce the training data. Several works also devote efforts to environment denoising \cite{lipsync3d_2021} and lip coherence \cite{stableface_2022} for the more stable performance.
The image-warping pipeline struggles to handle the background and face distortion issues, resulting in inauthentic synthetic results. In this paper, the proposed method is built on the template-editing pipeline to achieve better video synthesizing quality.

\noindent
\textbf{Style modeling in talking avatar.} Recently,  several studies \cite{mead_2020, evp_2021, wu_2021, ned_2022, eamm_2022, gavc_2022, styletalk_2023} have explored how to generate a more vivid avatar by modeling the facial talking style. MEAD \cite{mead_2020} collects an emotional talking face dataset and achieves coarse-grained emotion control. EVP \cite{evp_2021} utilizes this dataset and synthesizes more emotional dynamics by decoupling speech into different latent features. 
NED \cite{ned_2022} attempts to explore the style representation from the reference video, but it still needs labeled data for training. In the meantime, NED only manipulates the emotion and preserves the speech, which is not audio-driven. GC-AVT \cite{gavc_2022} and EAMM \cite{eamm_2022} also incorporate MEAD dataset, and they struggle to derive emotion factor from the raw images, which is not flexible for arbitrary subjects. 
The most-related work is StyleAvatar \cite{wu_2021}, which attempts to employ the manual features to define talking style and can capture the dynamic change of the facial expression. 
However, the manual features have limited representation ability for the expressive facial style. 
And StyleTalk \cite{styletalk_2023}, the latest image-warping based work, extracts speaking style from a reference video. However, it has limitations in preserving sufficient style information due to the use of a flat style code, and it fails to retain the original portrait identity.

\section{Method}
This paper proposes a variational style transfer model for vivid talking avatar generation.
Given a speech utterance and a video prompt as inputs, our method generates a vivid
avatar, which reenacts the mouth movements that are synchronized with the speech and takes on the facial style of the video prompt. The proposed VAST includes a style encoder, a variational style enhancer, and a hybrid decoder. The encoder and decoder realizes the zero-shot style transfer with the learned flat style space. The variational style enhancer equips the style space with complex dense distribution with the normalizing flow \cite{householderflow_2017} enhanced variational autoencoder (VAE) \cite{vae_2013, cvae_2015}. The image renderer is inspired by StableFace \cite{stableface_2022}.

To  be more specific, as shown in Fig. \ref{fig: overview}, the proposed method generates the vivid avatar with the following steps: 1) The input speech is represented as the speaker and language independent feature phonetic posteriorgram (PPG) \cite{ppg_2016} $\boldsymbol{A}=[\boldsymbol{a}_1, \ldots, \boldsymbol{a}_T]$, where $T$ is the frame length of the PPG sequence. 2) A parametric face model $\mathbf{\Phi}$ \cite{moai_2021} is adopted to achieve high-level fully disentangled facial parameters (\emph{e.g.}, identity, expression, and pose) from the input video prompt $\boldsymbol{I}^p=[\boldsymbol{I}^p_1, \ldots, \boldsymbol{I}^p_N]$, where $N$ is the frame length of the video prompt. 3) The extracted expression $\boldsymbol{X}=[\boldsymbol{x}_1, \ldots, \boldsymbol{x}_N] \in \mathbb{R}^{N \times 233}$ is sent into a style encoder $\mathbf{E}^{p}$ to obtain a compact style embedding $\boldsymbol{s}$. 4) A variational style enhancer $\mathbf{E}^{v}$ is followed to enrich the learned style space. 5) The hybrid decoder $\mathbf{D}$, including an autoregressive decoder $\mathbf{D}^{ar}$ and a non-autoregressive decoder $\mathbf{D}^{nar}$, then predicts expression $\hat{\boldsymbol{X}}$ that conforms with the speech and the facial style in the video prompt. 6) $\hat{\boldsymbol{X}}$ and other facial parameters of the template video are sent into the decoding procedure $\mathbf{\Phi}^{\prime}$ of the face model to generate the mesh sequence $\boldsymbol{M}$. 7) An image renderer $\mathbf{G}$ is finally adopted to synthesize photo-realistic video $\hat{\boldsymbol{I}}$.
 We will describe detailed designs of the proposed method in the following subsections.

\subsection{Style Encoder}
We firstly introduce how to obtain robust facial style representation from various video prompt. 
To reduce the input complexity and remove unnecessary information, the style representation is learned from the extracted expression sequence $\boldsymbol{X}$ instead of the raw video. The variable-length sequence $\boldsymbol{X}$ is passed through convolution and recurrent layers and then compressed as a fixed-length vector $\boldsymbol{s} \in \mathbb{R}^{d_{\boldsymbol{s}}}$. This structure is usually adopted in learning the embedding for the sequence data \cite{prosody_2018}. 
We consider this embedding as the style space, and sampling from this space will yield natural facial style. The facial style is compressed as a fixed-length vector other than a variable-length sequence, since the style in a short-length video sample hardly changes. This style encoder is also expected to perceive speech information and fuse it with the facial style, since the speech affects the presentation of facial style. Thus, we incorporate PPG into the encoder.


\subsection{Variational Style Enhancer}
Previous works \cite{nvp_2020, evp_2021, lipsync3d_2021, wu_2021} usually adopt a deterministic model with the mean square error (MSE) loss to predict expression from speech. This always leads to the mean movement prediction of the training data and decreases the expressiveness. To generate more expressive expression other than mean movements, we design a variational style enhancer to equip the style space with a complex distribution.
 We treat the style vector $\boldsymbol{s}$ as the latent variable $\boldsymbol{z}$ in VAE \cite{vae_2013, cvae_2015} framework with a learnable prior distribution $p(\boldsymbol{z} \mid \boldsymbol{A})$.
Specifically, we develop our model following the variational lower bound of the likelihood:
\begin{align}
    \label{eq:cvae}
    \ln p_{\theta}(\boldsymbol{X} \mid \boldsymbol{A})
    &\geq \mathbb{E}_{q_{\phi}(\boldsymbol{z} \mid \boldsymbol{X}, \boldsymbol{A})}[\ln p_{\theta}(\boldsymbol{X} \mid \boldsymbol{z}, \boldsymbol{A})] \\ \notag
    &-\mathrm{KL}(q_{\phi}(\boldsymbol{z} \mid \boldsymbol{X}, \boldsymbol{A}) \| p(\boldsymbol{z})),
\end{align}
where $\mathrm{KL}$ represents the Kullback-Leibler (KL)-divergence. To be more specific, $q_{\phi}(\boldsymbol{z} \mid \boldsymbol{X}, \boldsymbol{A})$ is the style encoder and $p_{\theta}(\boldsymbol{X} \mid \boldsymbol{z}, \boldsymbol{A})$ is the hybrid expression decoder. The usual choice of $q_{\phi}(\boldsymbol{z} \mid \boldsymbol{X}, \boldsymbol{A})$ is a diagonal covariance Gaussian $\mathcal{N}(\boldsymbol{z} \mid \boldsymbol{\mu}(\boldsymbol{X}, \boldsymbol{A}), \boldsymbol{\sigma}^2(\boldsymbol{X}, \boldsymbol{A})\mathbf{I})$.

\noindent
\textbf{Normalizing Flow Enhancement.} However, the diagonal posterior in the vanilla VAE is insufficient enough to match the true posterior \cite{householderflow_2017}.
In this paper, we introduce the normalizing flow to enrich the diagonal posterior. 
We try to pursue the full-covariance matrix instead of the diagonal matrix in the vanilla VAE by applying a series of invertible householder-flow ($\mathbf{HF}$) \cite{householderflow_2017} based transformations $\mathbf{H}^{(k)}$ on the initial latent variable $\boldsymbol{z}^{(0)}$. After these $K$ transformations, we can sample from a more flexible posterior $\boldsymbol{z}^{(K)}$.
The training loss for the variational style transfer model is:
\begin{align}
    \label{eq:loss_st}
    \mathcal{L}(\mathbf{E}^{p}, \mathbf{E}^{v}, \mathbf{D}) &= \operatorname{KL}(q_{\phi}({\boldsymbol{z}}^{(0)} \mid \boldsymbol{X}, \boldsymbol{A})|| p({\boldsymbol{z}}^{(K)})) \notag \\
    &-\mathbb{E}_{q_{\phi}({\boldsymbol{z}}^{(0)} \mid \boldsymbol{X}, \boldsymbol{A})}[\ln p_{\theta}(\boldsymbol{X} \mid {\boldsymbol{z}}^{(K)}, \boldsymbol{A}) \notag \\
    &-\sum_{k=1}^K \ln |\operatorname{det}{\frac{\partial \mathbf{H}^{(k)}}{\partial {\boldsymbol{z}}^{(k-1)}}}|]
\end{align}
The second term in Eq. \ref{eq:loss_st} refers to the MSE loss, and it also leads to the mean movement problem. To further alleviate this problem, we replace the MSE loss with an asymmetric reconstruction loss \cite{asymmetric_2017}, which helps escape the mean movement problem and generate more detailed and expressive movements. Detailed mathematical derivations of the loss function are provided in the \emph{Sup. Mat.}.

\subsection{Hybrid Facial Expression Decoder}
To robustly generate speech-synchronized expressions, we split the expression to speech-weakly-related parameters and speech-strongly-related parameters, and specially design the decoder architecture.
Previous methods \cite{nvp_2020, wu_2021, lipsync3d_2021} usually adopt a single model to predict the expressions of all dimensions. They neglect an important feature of human facial movements: the movements on some regions (\emph{e.g.}, mouth, chin, lip) have strong relationship with speech while movements in other regions (\emph{e.g.}, eyes, forehead) are weakly-related with speech. The former movements are vital for the lip synchronization, and the prediction should be accurate-enough. The latter movements are just spontaneous muscle movements that naturally inherit from previous time steps, and this prediction does not require high accuracy. On the contrary, a powerful network will regress the
outputs to a mean movement. For example, 90\% eye-related parameters are recorded as eye-open state in the training data, a powerful network trained on such data will always generate eye-open results. Therefore, a hybrid decoder is adopted to separately predict these two types of movements.


\noindent
\textbf{Facial Expression Categorization.}
Since not all expressions \cite{moai_2021} are defined with physical meaning of facial movements, we start by exploring which expressions are strongly and weakly correlated with speech. Such correlations are computed as the vertex offset around the mouth region. For each expression parameter defined by the face model \cite{moai_2021}, we record the maximum vertex offset caused by the parameter changing within a certain range. The larger offset it causes, the stronger correlation it has with speech. 
We empirically select an appropriate offset threshold to split these parameters into speech-weekly-related $\boldsymbol{x}^{wk} \in \mathbb{R}^{148}$ and speech-strongly-related parameters $\boldsymbol{x}^{st} \in \mathbb{R}^{85}$. 

\noindent
\textbf{Autoregressive Decoder.}
For $\boldsymbol{X}^{wk}$ (sequence of $\boldsymbol{x}^{wk}$), an autoregressive model $\mathbf{D}^{ar}$ with a simple architecture of long short-term memory network (LSTM) \cite{lstm_1997} is adopted since it is biased to learn the dependency of data itself across time. The autoregressive model also specializes in generation with weakly-related control signal \cite{wavenet_2016}. We employ PPG as the condition of this prediction, and PPG is masked with dropout \cite{dp_2014} to manually prevent overfitting the spurious correlation.
This prediction can be formulated as:
\begin{align}
    \hat{\boldsymbol{x}}^{wk}_{t} = \mathbf{D}^{ar}(\boldsymbol{x}^{wk}_{1:t-1}, \mathbf{DP}(\boldsymbol{a}_{1:t-1}), \boldsymbol{s}),
\end{align}
where $\mathbf{DP}$ is the dropout function.

\noindent
\textbf{Non-Autoregressive Decoder.}
For $\boldsymbol{X}^{st}$ (sequence of $\boldsymbol{x}^{st}$), a non-autoregressive model $\mathbf{D}^{nar}$ with the architecture of Transformer \cite{transformer_2017} is adopted for higher accuracy and greater semantic modeling ability. Several Transformer blocks are stacked to encode the input features of PPG and broadcast-repeated style embedding. Unlike other methods \cite{lipsync3d_2021, wu_2021, nvp_2020} predicting one-frame expression using multiple speech frames as input, we directly treat this prediction as a sequence-to-sequence task for better movement coherence:
\begin{align}
    \hat{\boldsymbol{X}}^{st} = \mathbf{D}^{nar}(\boldsymbol{A}, \boldsymbol{s}).
\end{align}
The design of hybrid decoder has no influence on the training loss and VAST is still optimized by Eq. \ref{eq:loss_st}. 

\subsection{Image Renderer}
The final module of our method is a photorealistic image renderer $\mathbf{G}$.  
The renderer takes the mesh images $\boldsymbol{M}$ decoded by the face model \cite{moai_2021} as input, and synthesizes portrait images $\hat{\boldsymbol{I}}$. The synthesized images match the shape and expression details of $\boldsymbol{M}$, and are inpainted with the same appearance of the template video. We adopt L1 pixel-wise loss and the perceptual VGG loss \cite{johnson2016perceptual} for image reconstruction:
\begin{equation}
    \mathcal{L}_{rec}(\mathbf{G}) = \|\hat{\boldsymbol{I}} - \boldsymbol{I}\|_1 + \mathbf{VGG}(\hat{\boldsymbol{I}}, \boldsymbol{I}),
\end{equation}
where $\boldsymbol{I}$ is the ground-truth images, and $\mathbf{VGG}(\hat{\boldsymbol{I}}, \boldsymbol{I})$ denotes the L2 loss of features extracted by VGG network~\cite{simonyan2014very}.

To improve the image quality and facial details, we employ the conditional generative adversarial (cGAN) loss \cite{cgan_2017} to train a discriminator $\mathbf{D}$ and the generator $\mathbf{G}$. The ground-truth images $\boldsymbol{I}$ with its corresponding mesh images $\boldsymbol{M}$ are annotated as the real sample pair, while the generated images $\hat{\boldsymbol{I}}$ and $\boldsymbol{M}$ are the fake. 
The loss functions can be written as:
\begin{align}
    \mathcal{L}_{cGAN}(\mathbf{G}, \mathbf{D}) &= \mathbb{E}_{\boldsymbol{M}, \boldsymbol{A}, \boldsymbol{I}}[\log \mathbf{D}(\boldsymbol{M}, \boldsymbol{I})] \\ \notag
    &+ \mathbb{E}_{\boldsymbol{M}, \boldsymbol{A}}[\log (1-\mathbf{D}(\boldsymbol{M}, \mathbf{G}(\boldsymbol{M}, \boldsymbol{A}))]
\end{align}
To improve the consistency across the consecutive frames, we adopt a U-Net structure as~\cite{nvp_2020} the basic backbone for $\mathbf{G}$ and enhance it with LSTM~\cite{lstm_1997} layer. Different from \cite{nvp_2020,lipsync3d_2021}, the speech feature $\boldsymbol{A}$ is also taken as input to improve the speech-lip synchronization in the synthesized video.

The generator and discriminator are optimized by:
\begin{align}
    \mathbf{G}^{*}, \mathbf{D} =\arg \min _{\mathbf{G}} \max _{\mathbf{D}} \mathcal{L}_{c G A N}(\mathbf{G}, \mathbf{D})+ \mathcal{L}_{rec}(\mathbf{G}).
\end{align}

\begin{figure*}[t]
    \centering
    \includegraphics[width=1.0\textwidth]{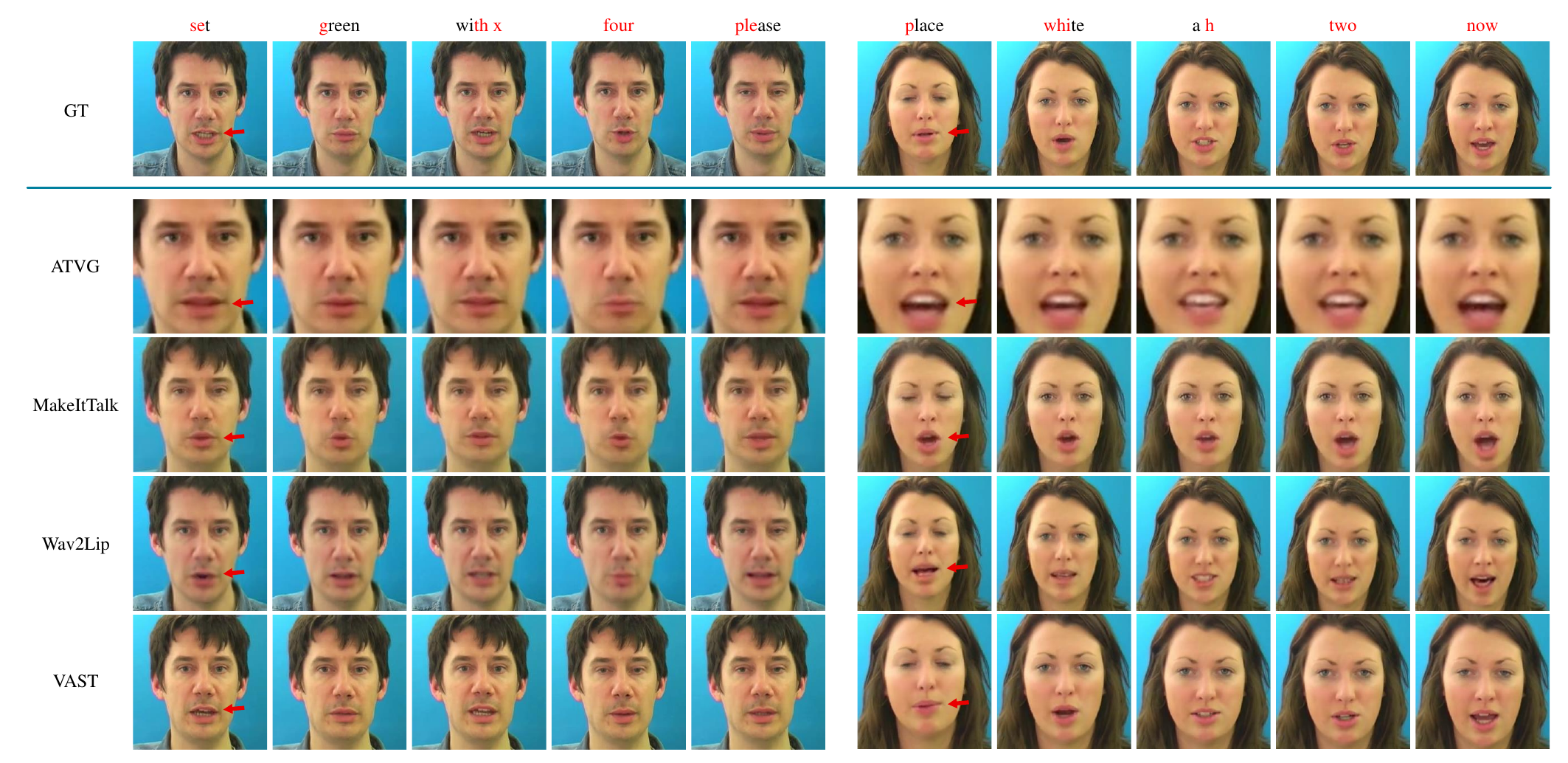} 
    \caption{Qualitative comparison results on GRID dataset. ATVG \cite{atvg} produces blurry images. MakeItTalk \cite{makeittalk_2020} and Wav2Lip \cite{wav2lip_2020} generate wrong lip movements in cases. The video prompt to provide style for our method is randomly selected from other videos of the avatars that are intended to be synthesized.}
    \label{fig:qualitative}
\end{figure*}
\begin{table*}[tbp]
  \small
  \centering
  \caption{Quantitative comparison resuls on GRID dataset. M-Sync: MOS for synchronization. M-Nat: MOS for naturalness.}
    \begin{tabular}{lcccccccccc}
    \toprule
    Method & CPBD$\uparrow$  & SSIM$\uparrow$  & FID$\downarrow$   & LMD$\downarrow$   & LMD-m$\downarrow$ & Sync-Dist$\downarrow$ & Sync-Conf$\uparrow$ & WER$\downarrow$ & M-Sync$\uparrow$ & M-Nat$\uparrow$\\ 
    \midrule
    GT    & 0.2650 & -      & -        & 0     & 0     & 6.959 & 7.039 & 0.067 & 4.65 & 4.72\\
    \hline
    ATVG~\cite{atvg}  & 0.068     & 0.830 & 56.4 & 2.738 & 2.707 & 8.031 & 5.434 & 0.633 & 3.69 & 3.55\\ 
    MakeItTalk~\cite{makeittalk_2020} & 0.1825 & 0.810 & 38.2  & 2.593 & 2.807 & 10.138 & 3.446 & 0.733 & 3.52 & 3.61\\ 
    Wav2Lip~\cite{wav2lip_2020} & 0.1174 & \textbf{0.946} & 22.6  & 1.616 & 1.760 & \textbf{6.472} & \textbf{8.038} & 0.500 & 3.84 & 3.82\\ 
    VAST & \textbf{0.2265} & 0.922 & \textbf{21.0}  & \textbf{1.444} & \textbf{1.479} & 6.656 & 7.571 & \textbf{0.233} & \textbf{4.53} & \textbf{4.38}\\ 
    \bottomrule
    \end{tabular}%
  \label{table: table1}%
\end{table*}%

\section{Experiments}
\subsection{Experimental Setup}
\noindent
\textbf{Dataset.} 
Four datasets are leveraged: 1) GRID dataset \cite{grid_2006}: a high-quality video corpus with the neutral talking style, recorded in laboratory conditions. We adopt the speaker s1 and s25 in this work. 2) Ted-HD \cite{wu_2021}: a 6-hour dataset collected from the Ted website. It contains 60 speakers with diverse and natural talking styles, but without any manual annotation. 3) Obama dataset: a single-speaker audio-visual dataset with a neutral broadcasting style. We collect 30-minute Obama Weekly Address videos following \cite{obama_2017}. 4) HDTF \cite{hdtf_2021}: a high-resolution in-the-wild video dataset, 10-minute video clips of two speakers are selected. 

\noindent
\textbf{Implementation Details.}
The original videos are downsampled to 25fps, cropped for talking faces \cite{facealign_2017} and resized to $256 \times 256$ pixels. The prepared images are sent into the parametric face model \cite{moai_2021} for facial parameters extraction. 
To obtain robust representation of speech PPG,
we pretrain an acoustic model \cite{deepspeech_2014} with a large amount of easily-available speech corpora. The speech is sampled at
16kHz. Filter banks of 80 dimensions with a sliding window of 40ms width and 20ms frame shift are utilized as the inputs of the acoustic model. The frame rates of PPG and facial expression are different in training, thus we adopt
the interpolation operation \cite{voca} to align the audio and visual
features to the same length $N$.
We adopt a two-step procedure to train the variational style transfer model and image renderer.  For the style transfer model, we adopt 80 percent of GRID and Ted-HD data for training with the loss defined in Eq. \ref{eq:loss_st}. These two datasets are mixed together without any identity label. For the renderer module, GRID, Obama and HDTF datasets are separately utilized for training the
person-specific avatars to visualize the vivid predicted expression.

\begin{figure*}[t]
    \centering
    \includegraphics[width=1.0\textwidth]{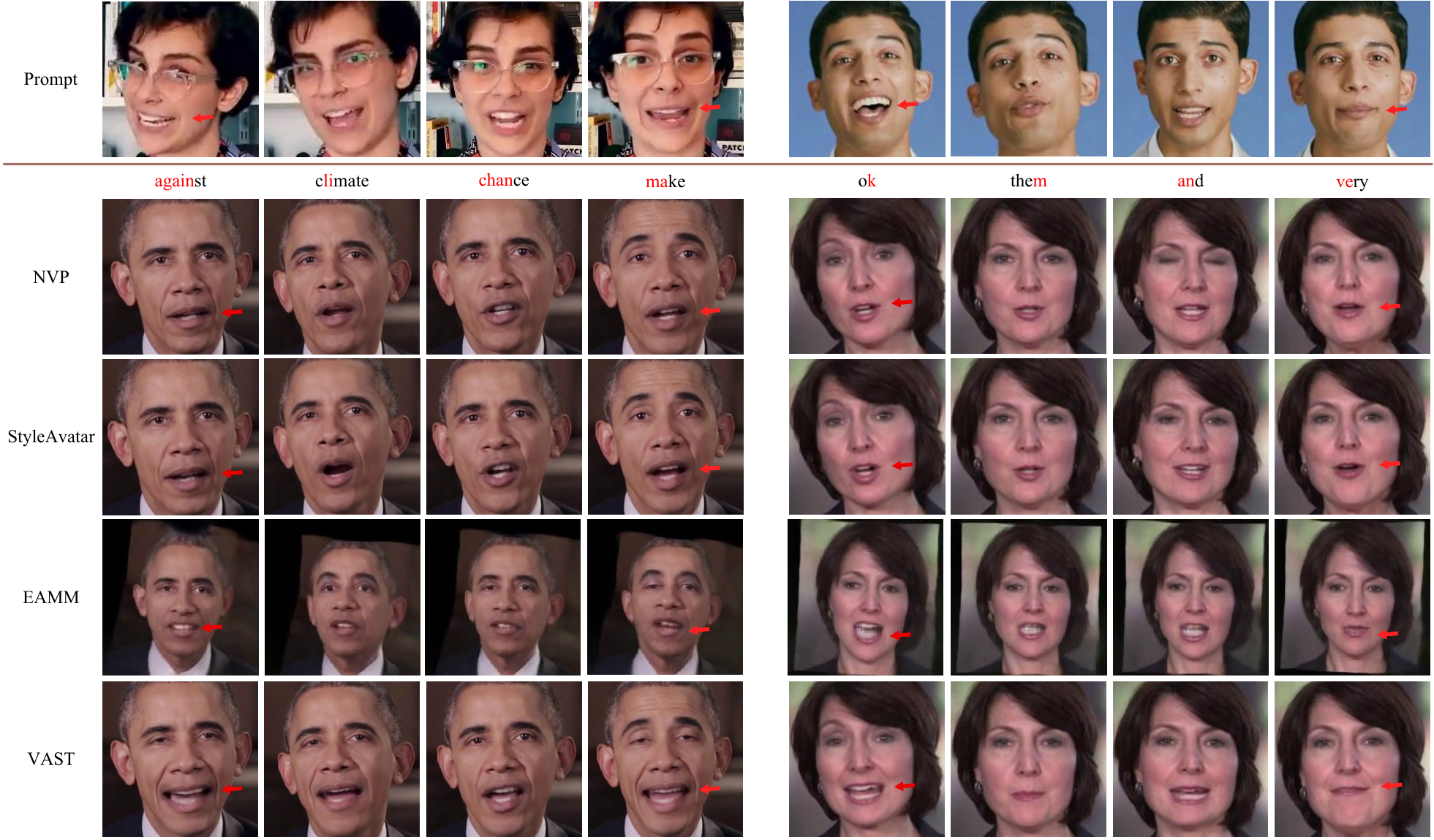} 
    \caption{Qualitative comparison results on Obama and HDTF datasets. The video prompts are sampled from the Ted-HD and HDTF datasets with styles like exciting-lecture (left female) and talk-show (right male). Our method is more like the facial style of the video prompts. Bigger mouth opening  at vowels (\emph{e.g.} /ei/ in make), tighter mouth shut at consonants (\emph{e.g.} /m/ in them), and even biting the lip (\emph{e.g.} /v/ in very) for our method.} 
    \label{fig:qualitative_style}
\end{figure*}

\begin{table}[tbp]
\footnotesize
\setlength\tabcolsep{4pt}
  \centering
  \caption{Expressiveness comparisons on Obama dataset. Dvt: the diversity metric \cite{bailando_2022}. all: expressions of all dimensions. st: speech-strongly-related expressions. Ori: expressions from the original Obama video. }
    \begin{tabular}{lcccccc}
    \toprule
    Metric & Ori    & NVP   & StyleAvatar & EAMM & VAST  & Prompt \\  
    \midrule
    Dvt-$\boldsymbol{X} \uparrow$ & 22.35 & 20.81 & 24.25 & 25.92 & \textbf{26.72} & 29.33 \\
    Dvt-$\boldsymbol{X}^{st} \uparrow$ & 27.49 & 23.96 & 30.42  & 31.12 & \textbf{36.32} & 36.15 \\
    MOS $\uparrow$ & - & 3.78 & 3.60 & 3.40 & \textbf{4.12} & -\\
    \bottomrule
    \end{tabular}%
  \label{tab:diversity}%
\end{table}%

\noindent
\textbf{Compared Methods.} 
Two tasks are performed in our experiments: the authenticity task, which evaluates lip synchronization and image quality, and the expressiveness task, which evaluates the effectiveness of transferred style onto the synthesized avatar.
For the authenticity evaluation, we compare our method with state-of-the-art audio-driven works which mainly focus on speech-lip synchronization.
\textbf{ATVG} \cite{atvg} generates video conditioned on the facial landmarks with dynamically adjustable pixel-wise loss and an attention mechanism. \textbf{MakeItTalk} \cite{makeittalk_2020} disentangles the content and speaker information to capture speaker-aware dynamics. \textbf{Wav2Lip} \cite{wav2lip_2020} accurately morphs the lip movements of arbitrary identities with an expert lip-sync discriminator.
For the expressiveness evaluation, we choose the code-available works that are related with the template-editing pipeline \cite{nvp_2020, wu_2021} and expressiveness modeling \cite{wu_2021, eamm_2022}.
\textbf{NVP} \cite{nvp_2020} presents a novel
audio-driven face reenactment approach that is generalized
among different audio sources. Another compared method
is \textbf{StyleAvatar} \cite{wu_2021}. Since the original StyleAvatar synthesizes images of poor quality, for fair comparison, we retain the style code design in StyleAvatar and use our renderer to generate full portrait images.
We also compare with \textbf{EAMM} \cite{eamm_2022}, which generates emotional talking faces by involving an emotion source video.

\begin{table*}[tbp]
\small
  \centering
  \caption{MOS and diversity \cite{bailando_2022} results for ablation study. The modules are removed from left to right step by step.}
    \begin{tabular}{lccccc}
    \toprule
    Metric & VAST & w/o $\mathbf{HF}$ & w/o $\mathbf{E}^{v}$ & w/o $\boldsymbol{s}$  & w/o $\mathbf{D}^{ar}$ \\
    \midrule
    Speech-Lip Sync $\uparrow$ & \textbf{3.89$\pm$0.21} & 3.82$\pm$0.19 & 3.86$\pm$0.22 & 3.49$\pm$0.21 & 3.42$\pm$0.22 \\
    Expressiveness \& Richness $\uparrow$ & \textbf{4.12$\pm$0.15} & 4.09$\pm$0.16 & 4.05$\pm$0.20 & 3.93$\pm$0.15 & 3.78$\pm$0.16 \\
    Overall Naturalness $\uparrow$ & 4.20$\pm$0.14 & 4.06$\pm$0.16 & \textbf{4.22$\pm$0.14} & 3.98$\pm$0.17 & 4.07$\pm$0.19 \\
    Diversity-$\boldsymbol{X} \uparrow$ & \textbf{26.72} & 26.53 & 25.80 & 21.06  & 20.81 \\
    \bottomrule
    \end{tabular}%
  \label{tab:MOS}%
\end{table*}

\noindent
\textbf{Objective Metrics.} We compute several objective metrics that have been widely adopted in previous works \cite{atvg, nvp_2020, lipsync3d_2021} to evaluate the image quality and lip synchronization. CPBD \cite{cpbd_2009} evaluates the sharpness and FID \cite{fid_2017} measures the realness of images. SSIM \cite{ssim} measure the reconstruction quality of images.  LMD denotes the average distance of all landmarks \cite{facealign_2017} on faces between the ground-truth and synthesized images, while LMD-m only calculates the mouth-region landmarks. To remove the head pose influence when calculating LMD, the Umeyama algorithm \cite{umeyama_1991} is introduced for landmark normalization. Sync-Dist and Sync-Conf \cite{lipnet_2016} are employed to score the speech-lip synchronization performance. WER (word error rate) \cite{lipnet_2016} measures the accuracy of lip reading from videos.

\noindent
\textbf{Subjective Metrics.} To evaluate the perceptual quality and expressiveness on generated videos, mean opinion score (MOS) tests   are conducted in this work. Fifteen participants with proficient English ability are invited in these tests, and they are asked to give a score from 1 (worst) to 5 (best) on the test videos towards different aspects: speech-lip synchronization, expressiveness, overall naturalness.

\subsection{Authenticity Evaluation}
\noindent
\textbf{Qualitative Comparison.} The qualitative comparison results with other methods is shown in Fig. \ref{fig:qualitative}. With the same speech input, we randomly select several frames synthesized by different methods and assign these frames with the words or phonemes that are being spoken. By zooming in on the images, we can find that other methods produce blur images, especially for the mouth and teeth regions. The proposed method produces sharp details, which are closer to the ground-truth images. The proposed method also generates highly speech-synchronized lip movements. When the avatar is speaking the voiceless consonants (\emph{e.g.} /p/ in place), the mouth can be shut tightly. When it comes to the vowels (\emph{e.g.} /ai/ in white, /e/ in set), the lip movement is more expressive.

\begin{figure}[tbp]
    \centering
    \includegraphics[width=0.5\textwidth]{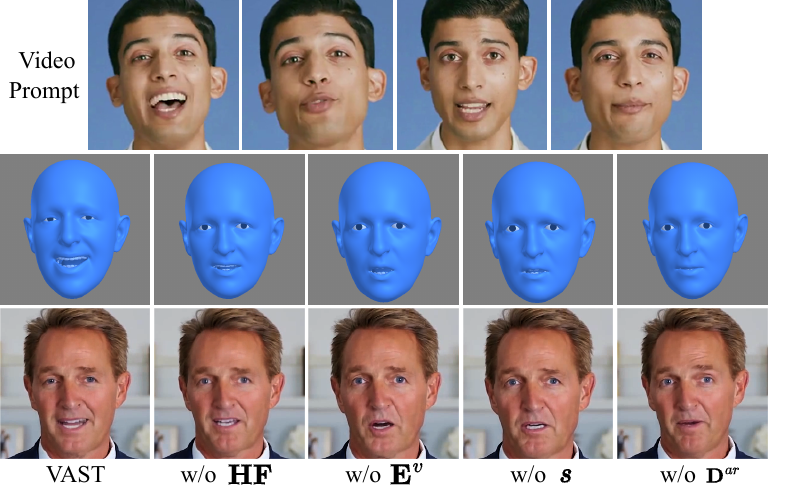}
    \caption{Qualitative result for ablation on HDTF dataset. The avatar is speaking /ei/ of pray. }
    \label{fig:ablation_qualitative}
\end{figure}

\noindent
\textbf{Quantitative Comparison.} The quantitative comparison results are presented in Table \ref{table: table1}. Compared with other methods, our method achieves competitive performance on image quality, although the renderer architecture is not specially designed.  In terms of speech-lip synchronization, it can be observed that Sync-Dist and Sync-Conf by our method are close to the ground-truth. It should be noticed that Wav2Lip directly adopts SyncNet \cite{lipnet_2016} as its discriminator, thus gains better results on this metric by design. Our method obtains the lowest LMD and LMD-m, which proves that our method generates lip movements of the highest accuracy. These two metrics are actually more reliable and objectively fair, since they simply calculate the landmark mean error. The lowest WER result achieved by our method also proves the accurate lip movement generation. 

\noindent
\textbf{Subjective Comparison.} The subjective comparison results are presented in Table \ref{table: table1}. The proposed VAST has achieved the best performance on both lip synchronization and overall naturalness.

\subsection{Expressiveness Evaluation}
\noindent
\textbf{Qualitative Comparison.} 
The qualitative comparison results are shown in Fig. \ref{fig:qualitative_style}. 
Our method is able to generate more exaggerated and expressive facial movements. The avatar can grin and open the mouth with a larger amplitude. The wrinkles around the mouth region are deeper, which indicates that the muscle movements are more powerful and the expression around the cheek are richer. NVP  opens the mouth only in a small range. StyleAvatar opens the mouth slightly bigger, since it actually introduces a bias signal (mean and variance of the reference video) to the generation process. EAMM generates exaggerated expressions but bad visual results. The proposed approach can produce more vivid movements compared with other methods.

\noindent
\textbf{Quantitative Comparison.} We adopt the diversity metric to estimate the expressiveness of facial movements, which has been widely adopted in the gesture and dance generation field \cite{dance2music_2019, dancerevo_2020, ast_2021, bailando_2022}. We compute the average Euclidean distance of the generated expression sequences. This metric reflects the variation of the sequence \cite{dancerevo_2020}. The diversity of all-dimension expressions $\boldsymbol{X}$ and speech-strongly-related expressions $\boldsymbol{X}^{st}$ are respectively computed. As illustrated in Table \ref{tab:diversity}, NVP obtains the lowest diversity, since it directly predicts facial expression from audio and is optimized by the MSE loss. Such a deterministic model always regresses to an average output, leading to deficient expressiveness. StyleAvatar achieves higher diversity due to the residual information it introduces with the style codes. Our method significantly outperforms the baselines, and the speech-related diversity has been improved greatly. 

\noindent
\textbf{Subjective Comparison.}
The subjective comparison on expressiveness is presented in the last row of Table \ref{tab:diversity}. The proposed VAST has achieved the best performance on expressiveness compared with state-of-the-art methods, with at least 14.4$\%$ relative improvement.

\begin{figure}[tbp]
  \centering
  \begin{subfigure}{0.48\linewidth}
    \centering
    \includegraphics[scale=0.3]{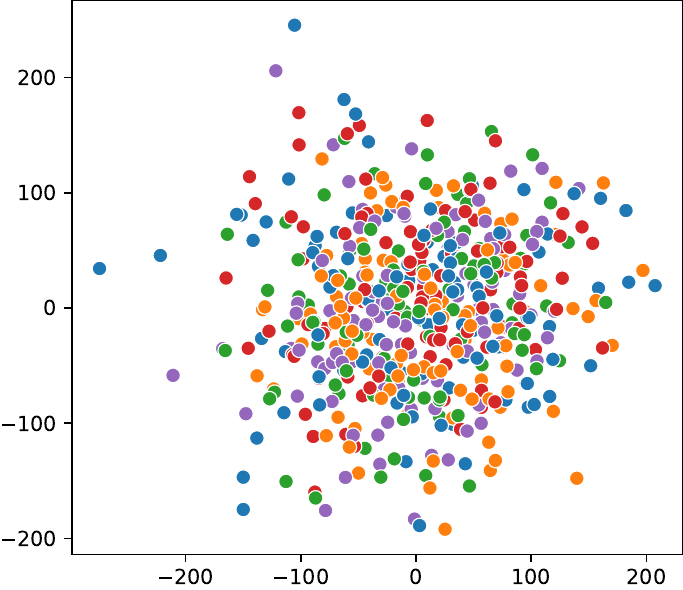}
    \caption{w/o $\mathbf{E}^{v}$.}
    \label{fig:tsne_novae}
  \end{subfigure}
  \centering
  \begin{subfigure}{0.48\linewidth}
    \centering
    \includegraphics[scale=0.3]{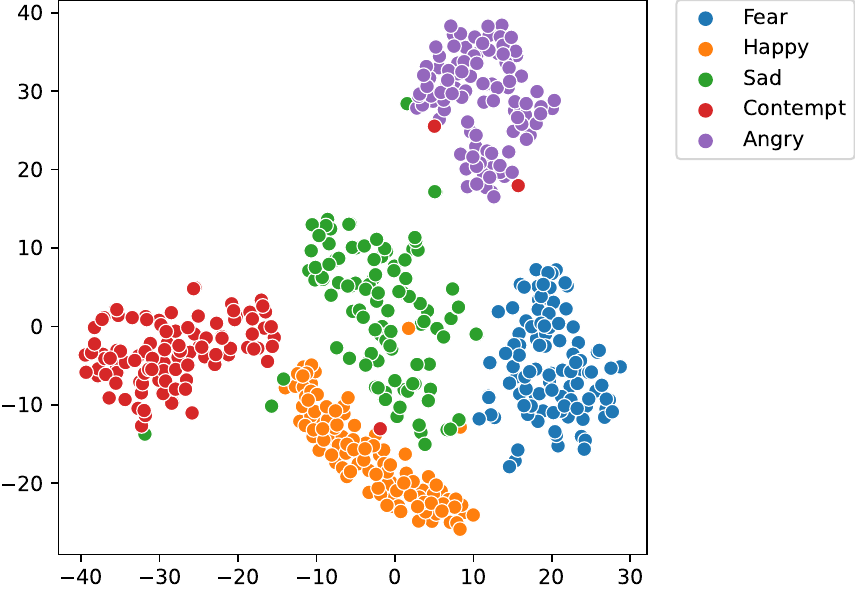}
    \caption{w/ $\mathbf{E}^{v}$.}
    \label{fig:tsne_wvae}
  \end{subfigure}
  \caption{Visualization of the learned style embedding.}
  \label{fig:tsne}
\end{figure}

\subsection{Ablation Study}
\noindent
\textbf{Qualitative Result.} We present the qualitative result for ablation in Fig. \ref{fig:ablation_qualitative}. It can be observed that VAST with complete designed modules achieves accurate lip movements when speaking /ei/, and also transfers the most similar facial style from the video prompt. The mesh results in the second row are easier to observe the advantage of our method.

\noindent
\textbf{Subjective Evaluation \& Diversity Metric.}
To verify the effectiveness of the designed modules, we extensively conduct MOS test on the synthesized videos. We randomly select 50 videos for test, 10 videos for each category. 
The MOS results are shown in Table \ref{tab:MOS}. The proposed method achieves the highest scores in the terms of speech-lip synchronization and expressiveness \& richness. Without the variational style enhancer $\mathbf{E}^{v}$, the expressiveness has a significant drop. The normalizing flow module $\mathbf{HF}$ proves to perceptibly benefit the VAE posterior, which is consistent with the theoretical design \cite{householderflow_2017}. The style embedding $\boldsymbol{s}$ proves to be crucial for lip sync, since it provides the residual information that cannot be predicted from the speech input. The absence of the autoregressive decoder $\mathbf{D}^{ar}$ also leads to the performance  drop. This indicates that categorizing the facial expressions and introducing the specific inductive bias for $\boldsymbol{X}^{wk}$ are practical and effective. We also notice that these modules have less influence on the overall naturalness because the template-editing pipeline has guaranteed the naturalness and realness of synthesized videos.

\noindent
\textbf{Visualization of Style Space.} Since the designed variational style transfer model tends to learn a more meaningful latent space, we visualize the latent embeddings. MEAD \cite{mead_2020} dataset is introduced here to provide emotion labels, and five emotions are selected.
We extract expressions from MEAD and obtain the style embeddings generated by the style encoder and variational style enhancer. The t-SNE algorithm \cite{tsne_2008} is utilized to reduce the data dimension and plot it onto the 2D plane. As shown in Fig. \ref{fig:tsne}, the embeddings obtained without the variational style enhancer are mingled in chaos, while the variational module contributes to more organized clusters of different emotions. It demonstrates that introducing variational style enhancer results in meaningful representation learning. Although the proposed model has never been trained on any emotion-labeled data, it explores and summarizes the semantic relationships between facial expressions and emotions in an unsupervised manner, and generalizes to MEAD dataset.
\begin{figure}[tbp]
    \centering
    \includegraphics[width=0.47\textwidth]{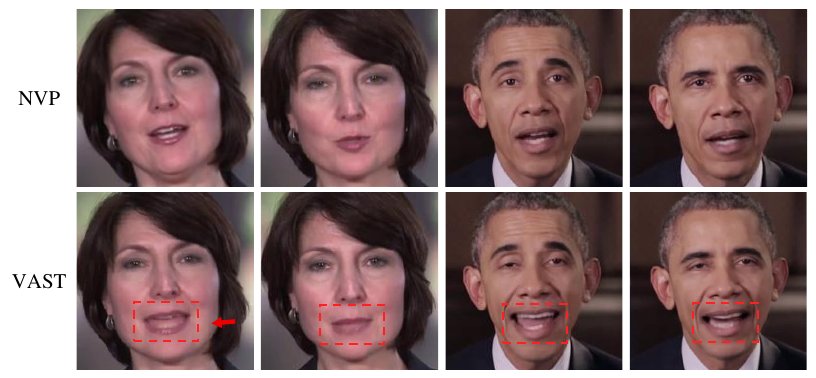} 
    \caption{Failed cases of the image renderer when the style is too exaggerated.}
    \label{fig:badcase}
\end{figure}

\subsection{Limitations}
Although the proposed method can transfer arbitrary facial style to the neutral avatar, there exists bad cases. 
As shown in Fig. \ref{fig:badcase}, the avatar synthesized by our method has noticeable artifacts around the mouth region. Since the renderer in VAST is not specially designed and the data for training the renderer lacks expressiveness, our method may synthesize some blur results when the video prompt has too exaggerated style. A renderer with more powerful structure trained on a wider range of data may solve this issue.

\section{Conclusion}
This paper proposes VAST, a zero-shot facial style transfer method for generating vivid talking avatars. VAST learns a discriminative and meaningful facial style representation from arbitrary video prompts using a style encoder and a variational style enhancer. A hybrid facial expression decoder is employed to transfer this representation onto a neutral avatar with high authenticity and rich expressiveness. Qualitative and quantitative results demonstrate the superiority of VAST over the state-of-the-art methods.
 

\vspace{0.25cm}
\textbf{Acknowledgement.} This work is supported by National Natural Science Foundation of China (62076144), Shenzhen Key Laboratory of next generation interactive media innovative technology (ZDSYS20210623092001004) as well as Shenzhen Science and Technology Program (WDZC20220816140515001, JCYJ20220818101014030).

{\small
\bibliographystyle{ieee_fullname}
\bibliography{vast}

\begin{thebibliography}{10}\itemsep=-1pt

\bibitem{facealign_2017}
Adrian Bulat and Georgios Tzimiropoulos.
\newblock How far are we from solving the {2D} \& {3D} face alignment problem?
  ({And} a dataset of 230,000 {3D} facial landmarks).
\newblock In {\em Proceedings of the IEEE International Conference on Computer
  Vision (ICCV)}, pages 1021--1030, 2017.

\bibitem{atvg}
Lele Chen, Ross~K Maddox, Zhiyao Duan, and Chenliang Xu.
\newblock Hierarchical cross-modal talking face generation with dynamic
  pixel-wise loss.
\newblock In {\em Proceedings of the IEEE Conference on Computer Vision and
  Pattern Recognition (CVPR)}, pages 7832--7841, 2019.

\bibitem{duallip_2020}
Weicong Chen, Xu Tan, Yingce Xia, Tao Qin, Yu Wang, and Tie-Yan Liu.
\newblock Duallip: A system for joint lip reading and generation.
\newblock In {\em Proceedings of the ACM International Conference on Multimedia
  (ACMMM)}, page 1985–1993, 2020.

\bibitem{lipnet_2016}
Joon~Son Chung and Andrew Zisserman.
\newblock Lip reading in the wild.
\newblock In {\em Proceedings of the Asian Conference on Computer Vision
  (ACCV)}, pages 87--103, 2016.

\bibitem{grid_2006}
Martin Cooke, Jon Barker, Stuart Cunningham, and Xu Shao.
\newblock {An audio-visual corpus for speech perception and automatic speech
  recognition}.
\newblock {\em Acoustical Society of America Journal (JASA)}, 120(5):2421,
  2006.

\bibitem{voca}
Daniel Cudeiro, Timo Bolkart, Cassidy Laidlaw, Anurag Ranjan, and Michael~J.
  Black.
\newblock Capture, learning, and synthesis of {3D} speaking styles.
\newblock In {\em Proceedings of the IEEE Conference on Computer Vision and
  Pattern Recognition (CVPR)}, pages 10093--10103, 2019.

\bibitem{deng_2019}
Yu Deng, Jiaolong Yang, Sicheng Xu, Dong Chen, Yunde Jia, and Xin Tong.
\newblock Accurate {3D} face reconstruction with weakly-supervised learning:
  From single image to image set.
\newblock In {\em Proceedings of the IEEE Conference on Computer Vision and
  Pattern Recognition Workshops (CVPRW)}, pages 285--295, 2019.

\bibitem{deepspeech_2014}
Awni Hannun, Carl Case, Jared Casper, Bryan Catanzaro, Greg Diamos, Erich
  Elsen, Ryan Prenger, Sanjeev Satheesh, Shubho Sengupta, Adam Coates, and
  Andrew~Y. Ng.
\newblock {Deep Speech}: Scaling up end-to-end speech recognition.
\newblock {\em arXiv preprint arXiv:1412.5567}, 2014.

\bibitem{fid_2017}
Martin Heusel, Hubert Ramsauer, Thomas Unterthiner, Bernhard Nessler, and Sepp
  Hochreiter.
\newblock {GANs} trained by a two time-scale update rule converge to a local
  nash equilibrium.
\newblock In {\em Proceedings of the International Conference on Neural
  Information Processing Systems (NIPS)}, page 6629–6640, 2017.

\bibitem{lstm_1997}
Sepp Hochreiter and Jürgen Schmidhuber.
\newblock Long short-term memory.
\newblock {\em Neural Computation}, 9(8):1735--1780, 1997.

\bibitem{dancerevo_2020}
Ruozi Huang, Huang Hu, Wei Wu, Kei Sawada, Mi Zhang, and Daxin Jiang.
\newblock {Dance Revolution}: Long-term dance generation with music via
  curriculum learning.
\newblock In {\em Proceedings of the International Conference on Learning
  Representations (ICLR)}, 2020.

\bibitem{cgan_2017}
Phillip Isola, Jun-Yan Zhu, Tinghui Zhou, and Alexei~A. Efros.
\newblock Image-to-image translation with conditional adversarial networks.
\newblock In {\em Proceedings of the IEEE Conference on Computer Vision and
  Pattern Recognition (CVPR)}, pages 5967--5976, 2017.

\bibitem{eamm_2022}
Xinya Ji, Hang Zhou, Kaisiyuan Wang, Qianyi Wu, Wayne Wu, Feng Xu, and Xun Cao.
\newblock {EAMM}: One-shot emotional talking face via audio-based emotion-aware
  motion model.
\newblock In {\em Proceedings of the ACM Special Interest Group on Computer
  Graphics and Interactive Techniques (SIGGRAPH)}, page~10, 2022.

\bibitem{evp_2021}
Xinya Ji, Hang Zhou, Kaisiyuan Wang, Wayne Wu, Chen~Change Loy, Xun Cao, and
  Feng Xu.
\newblock Audio-driven emotional video portraits.
\newblock In {\em Proceedings of the IEEE Conference on Computer Vision and
  Pattern Recognition (CVPR)}, pages 14080--14089, 2021.

\bibitem{johnson2016perceptual}
Justin Johnson, Alexandre Alahi, and Li Fei-Fei.
\newblock Perceptual losses for real-time style transfer and super-resolution.
\newblock In {\em Proceedings of the European Conference on Computer Vision
  (ECCV)}, pages 694--711, 2016.

\bibitem{vae_2013}
Diederik~P Kingma and Max Welling.
\newblock Auto-encoding variational {Bayes}.
\newblock {\em arXiv preprint arXiv:1312.6114}, 2022.

\bibitem{obamanet_2018}
Rithesh Kumar, Jose Sotelo, Kundan Kumar, Alexandre de Brebisson, and Yoshua
  Bengio.
\newblock {ObamaNet}: Photo-realistic lip-sync from text.
\newblock {\em arXiv preprint arXiv:1801.01442}, 2017.

\bibitem{lipsync3d_2021}
Avisek Lahiri, Vivek Kwatra, Christian Frueh, John Lewis, and Chris Bregler.
\newblock {LipSync3D}: Data-efficient learning of personalized {3D} talking
  faces from video using pose and lighting normalization.
\newblock In {\em Proceedings of the IEEE Conference on Computer Vision and
  Pattern Recognition (CVPR)}, pages 2755--2764, 2021.

\bibitem{dance2music_2019}
Hsin-Ying Lee, Xiaodong Yang, Ming-Yu Liu, Ting-Chun Wang, Yu-Ding Lu,
  Ming-Hsuan Yang, and Jan Kautz.
\newblock Dancing to music.
\newblock In {\em Proceedings of the International Conference on Neural
  Information Processing Systems (NIPS)}, volume~32, 2019.

\bibitem{ast_2021}
Ruilong Li, Shan Yang, David~A. Ross, and Angjoo Kanazawa.
\newblock {AI Choreographer}: Music conditioned {3D} dance generation with
  {AIST++}.
\newblock In {\em Proceedings of the IEEE International Conference on Computer
  Vision (ICCV)}, pages 13381--13392, 2021.

\bibitem{gavc_2022}
Borong Liang, Yan Pan, Zhizhi Guo, Hang Zhou, Zhibin Hong, Xiaoguang Han, Junyu
  Han, Jingtuo Liu, Errui Ding, and Jingdong Wang.
\newblock Expressive talking head generation with granular audio-visual
  control.
\newblock In {\em Proceedings of the IEEE Conference on Computer Vision and
  Pattern Recognition (CVPR)}, pages 3387--3396, 2022.

\bibitem{stableface_2022}
Jun Ling, Xu Tan, Liyang Chen, Runnan Li, Yuchao Zhang, Sheng Zhao, and Li
  Song.
\newblock {StableFace}: Analyzing and improving motion stability for talking
  face generation.
\newblock {\em arXiv preprint arXiv:2208.13717}, 2022.

\bibitem{livesp_2021}
Yuanxun Lu, Jinxiang Chai, and Xun Cao.
\newblock {Live Speech Portraits}: Real-time photorealistic talking-head
  animation.
\newblock {\em ACM Transactions on Graphics (TOG)}, 40(6):17, 2021.

\bibitem{styletalk_2023}
Yifeng Ma, Suzhen Wang, Zhipeng Hu, Changjie Fan, Tangjie Lv, Yu Ding, Zhidong
  Deng, and Xin Yu.
\newblock {StyleTalk}: One-shot talking head generation with controllable
  speaking styles.
\newblock In {\em Proceedings of the Association for the Advancement of
  Artificial Intelligence Conference (AAAI)}, volume~37, pages 1896--1904,
  2023.

\bibitem{cpbd_2009}
Niranjan~D. Narvekar and Lina~J. Karam.
\newblock A no-reference perceptual image sharpness metric based on a
  cumulative probability of blur detection.
\newblock In {\em Proceedings of the International Workshop on Quality of
  Multimedia Experience}, pages 87--91, 2009.

\bibitem{ned_2022}
Foivos Paraperas~Papantoniou, Panagiotis~P. Filntisis, Petros Maragos, and
  Anastasios Roussos.
\newblock {Neural Emotion Director}: Speech-preserving semantic control of
  facial expressions in ``in-the-wild'' videos.
\newblock In {\em Proceedings of the IEEE Conference on Computer Vision and
  Pattern Recognition (CVPR)}, pages 18759--18768, 2022.

\bibitem{synctalkingface_2022}
Se~Jin Park, Minsu Kim, Joanna Hong, Jeongsoo Choi, and Yong~Man Ro.
\newblock {SyncTalkFace}: Talking face generation with precise lip-syncing via
  audio-lip memory.
\newblock In {\em Proceedings of the Association for the Advancement of
  Artificial Intelligence Conference (AAAI)}, volume~36, pages 2062--2070,
  2022.

\bibitem{wav2lip_2020}
K~R Prajwal, Rudrabha Mukhopadhyay, Vinay~P. Namboodiri, and C.V. Jawahar.
\newblock A lip sync expert is all you need for speech to lip generation in the
  wild.
\newblock In {\em Proceedings of the ACM International Conference on Multimedia
  (ACMMM)}, page 484–492, 2020.

\bibitem{vi_2015}
Danilo Rezende and Shakir Mohamed.
\newblock Variational inference with normalizing flows.
\newblock In {\em Proceedings of the International Conference on Machine
  Learning (ICML)}, volume~37, pages 1530--1538, 2015.

\bibitem{simonyan2014very}
Karen Simonyan and Andrew Zisserman.
\newblock Very deep convolutional networks for large-scale image recognition.
\newblock In {\em Proceedings of the International Conference on Learning
  Representations (ICLR)}, 2015.

\bibitem{bailando_2022}
Li Siyao, Weijiang Yu, Tianpei Gu, Chunze Lin, Quan Wang, Chen Qian,
  Chen~Change Loy, and Ziwei Liu.
\newblock Bailando: {3D} dance generation by actor-critic {GPT} with
  choreographic memory.
\newblock In {\em Proceedings of the IEEE Conference on Computer Vision and
  Pattern Recognition (CVPR)}, pages 11050--11059, 2022.

\bibitem{prosody_2018}
RJ Skerry-Ryan, Eric Battenberg, Ying Xiao, Yuxuan Wang, Daisy Stanton, Joel
  Shor, Ron Weiss, Rob Clark, and Rif~A. Saurous.
\newblock Towards end-to-end prosody transfer for expressive speech synthesis
  with {Tacotron}.
\newblock In {\em Proceedings of the International Conference on Machine
  Learning (ICML)}, volume~80, pages 4693--4702, 2018.

\bibitem{cvae_2015}
Kihyuk Sohn, Honglak Lee, and Xinchen Yan.
\newblock Learning structured output representation using deep conditional
  generative models.
\newblock In {\em Proceedings of the International Conference on Neural
  Information Processing Systems (NIPS)}, volume~28, page 3483–3491, 2015.

\bibitem{makeittalk_2020}
Linsen Song, Wayne Wu, Chen Qian, Ran He, and Chen~Change Loy.
\newblock {Everybody’s Talkin’}: Let me talk as you want.
\newblock {\em IEEE Transactions on Information Forensics and Security (TIFS)},
  17:585--598, 2022.

\bibitem{dp_2014}
Nitish Srivastava, Geoffrey Hinton, Alex Krizhevsky, Ilya Sutskever, and Ruslan
  Salakhutdinov.
\newblock Dropout: A simple way to prevent neural networks from overfitting.
\newblock {\em Journal of Machine Learning Research}, 15(56):1929--1958, 2014.

\bibitem{ppg_2016}
Lifa Sun, Kun Li, Hao Wang, Shiyin Kang, and Helen Meng.
\newblock Phonetic posteriorgrams for many-to-one voice conversion without
  parallel data training.
\newblock In {\em Proceedings of the IEEE International Conference on
  Multimedia and Expo (ICME)}, pages 1--6, 2016.

\bibitem{obama_2017}
Supasorn Suwajanakorn, Steven~M. Seitz, and Ira Kemelmacher-Shlizerman.
\newblock Synthesizing {Obama}: Learning lip sync from audio.
\newblock {\em ACM Transactions on Graphics (TOG)}, 36(4):13, 2017.

\bibitem{nvp_2020}
Justus Thies, Mohamed Elgharib, Ayush Tewari, Christian Theobalt, and Matthias
  Nie\ss{}ner.
\newblock Neural voice puppetry: Audio-driven facial reenactment.
\newblock In {\em Proceedings of the European Conference on Computer Vision
  (ECCV)}, page 716–731, 2020.

\bibitem{householderflow_2017}
Jakub~M Tomczak and Max Welling.
\newblock Improving variational auto-encoders using householder flow.
\newblock In {\em Proceedings of the International Conference on Neural
  Information Processing Systems Workshops}, 2016.

\bibitem{asymmetric_2017}
Anh Tuan~Tran, Tal Hassner, Iacopo Masi, and Gerard Medioni.
\newblock Regressing robust and discriminative {3D} morphable models with a
  very deep neural network.
\newblock In {\em Proceedings of the IEEE Conference on Computer Vision and
  Pattern Recognition (CVPR)}, pages 1493--1502, 2017.

\bibitem{umeyama_1991}
S. Umeyama.
\newblock Least-squares estimation of transformation parameters between two
  point patterns.
\newblock {\em IEEE Transactions on Pattern Analysis and Machine Intelligence},
  13(4):376--380, 1991.

\bibitem{wavenet_2016}
Aäron van~den Oord, Sander Dieleman, Heiga Zen, Karen Simonyan, Oriol Vinyals,
  Alex Graves, Nal Kalchbrenner, Andrew Senior, and Koray Kavukcuoglu.
\newblock {WaveNet}: A generative model for raw audio.
\newblock In {\em Proceedings of the 9th ISCA Speech Synthesis Workshop}, page
  125, 2016.

\bibitem{tsne_2008}
Laurens Van~der Maaten and Geoffrey Hinton.
\newblock Visualizing data using {t-SNE}.
\newblock {\em Journal of machine learning research (JMLR)}, 9(11):2579--2605,
  2008.

\bibitem{transformer_2017}
Ashish Vaswani, Noam Shazeer, Niki Parmar, Jakob Uszkoreit, Llion Jones,
  Aidan~N Gomez, \L~ukasz Kaiser, and Illia Polosukhin.
\newblock Attention is all you need.
\newblock In {\em Proceedings of the International Conference on Neural
  Information Processing Systems (NIPS)}, volume~30, page 6000–6010, 2017.

\bibitem{mead_2020}
Kaisiyuan Wang, Qianyi Wu, Linsen Song, Zhuoqian Yang, Wayne Wu, Chen Qian, Ran
  He, Yu Qiao, and Chen~Change Loy.
\newblock {MEAD}: A large-scale audio-visual dataset for emotional talking-face
  generation.
\newblock In {\em Proceedings of the European Conference on Computer Vision
  (ECCV)}, page 700–717, 2020.

\bibitem{avct_2022}
Suzhen Wang, Lincheng Li, Yu Ding, and Xin Yu.
\newblock One-shot talking face generation from single-speaker audio-visual
  correlation learning.
\newblock In {\em Proceedings of the Association for the Advancement of
  Artificial Intelligence Conference (AAAI)}, volume~36, pages 2531--2539,
  2022.

\bibitem{ssim}
Zhou Wang, A.C. Bovik, H.R. Sheikh, and E.P. Simoncelli.
\newblock Image quality assessment: From error visibility to structural
  similarity.
\newblock {\em IEEE Transactions on Image Processing (TIP)}, 13(4):600--612,
  2004.

\bibitem{dvp_2020}
Xin Wen, Miao Wang, Christian Richardt, Ze-Yin Chen, and Shi-Min Hu.
\newblock Photo-realistic audio-driven video portraits.
\newblock {\em IEEE Transactions on Visualization and Computer Graphics
  (TVCG)}, 26(12):3457--3466, 2020.

\bibitem{moai_2021}
Erroll Wood, Tadas Baltru\v{s}aitis, Charlie Hewitt, Sebastian Dziadzio,
  Thomas~J. Cashman, and Jamie Shotton.
\newblock Fake it till you make it: Face analysis in the wild using synthetic
  data alone.
\newblock In {\em Proceedings of the IEEE International Conference on Computer
  Vision (ICCV)}, pages 3681--3691, 2021.

\bibitem{wu_2021}
Haozhe Wu, Jia Jia, Haoyu Wang, Yishun Dou, Chao Duan, and Qingshan Deng.
\newblock Imitating arbitrary talking style for realistic audio-driven talking
  face synthesis.
\newblock In {\em Proceedings of the ACM International Conference on Multimedia
  (ACMMM)}, page 1478–1486, 2021.

\bibitem{ye_2023}
Zhenhui Ye, Ziyue Jiang, Yi Ren, Jinglin Liu, Jinzheng He, and Zhou Zhao.
\newblock {GeneFace}: Generalized and high-fidelity audio-driven {3D} talking
  face synthesis.
\newblock In {\em Proceedings of the International Conference on Learning
  Representations (ICLR)}, 2023.

\bibitem{zhang2021facial}
Chenxu Zhang, Yifan Zhao, Yifei Huang, Ming Zeng, Saifeng Ni, Madhukar
  Budagavi, and Xiaohu Guo.
\newblock Facial: Synthesizing dynamic talking face with implicit attribute
  learning.
\newblock In {\em Proceedings of the IEEE International Conference on Computer
  Vision (ICCV)}, pages 3867--3876, 2021.

\bibitem{hdtf_2021}
Zhimeng Zhang, Lincheng Li, Yu Ding, and Changjie Fan.
\newblock Flow-guided one-shot talking face generation with a high-resolution
  audio-visual dataset.
\newblock In {\em Proceedings of the IEEE Conference on Computer Vision and
  Pattern Recognition (CVPR)}, pages 3661--3670, 2021.

\bibitem{zhou_2021}
Hang Zhou, Yasheng Sun, Wayne Wu, Chen~Change Loy, Xiaogang Wang, and Ziwei
  Liu.
\newblock Pose-controllable talking face generation by implicitly modularized
  audio-visual representation.
\newblock In {\em Proceedings of the IEEE Conference on Computer Vision and
  Pattern Recognition (CVPR)}, pages 4174--4184, 2021.

\end{thebibliography}
}

\newpage
\newpage
\setcounter{section}{0} 
\renewcommand{\thesection}{\Alph{section}}
\section{Variational Style Enhancer}
One of the key contributions of our method is that we propose the variational style enhancer, which enhances the
style space to be highly expressive and meaningful. Without this enhancer, the style space learned by the style encoder and hybrid decoder is flat. The variational style enhancer is based on variational autoencoder \cite{vae_2013, cvae_2015} and normalizing flow \cite{householderflow_2017}. The training of this enhancer can be considered as the reconstruction of facial expressions, which is conditional on the speech. If we remove the normalizing flow module, the loss function thus becomes:
\begin{align}
    \label{eq:cvae_supp}
    \ln p_{\theta}(\boldsymbol{X} \mid \boldsymbol{A})
    &\geq \mathbb{E}_{q_{\phi}(\boldsymbol{z} \mid \boldsymbol{X}, \boldsymbol{A})}[\ln p_{\theta}(\boldsymbol{X} \mid \boldsymbol{z}, \boldsymbol{A})] \\ \notag
    &-\mathrm{KL}(q_{\phi}(\boldsymbol{z} \mid \boldsymbol{X}, \boldsymbol{A}) \| p(\boldsymbol{z} | \boldsymbol{A})),
\end{align}
where $\boldsymbol{X}$ is the facial expression sequence, and $\boldsymbol{A}$ is the corresponding phonetic posteriorgram (PPG) \cite{ppg_2016} sequence. Since the latent variable $\boldsymbol{z}$ can be considered to be independent with $\boldsymbol{A}$, Eq. \ref{eq:cvae_supp} is further defined as
\begin{align}
    \label{eq:cvae2_supp}
    \ln p_{\theta}(\boldsymbol{X} \mid \boldsymbol{A})
    &\geq \mathbb{E}_{q_{\phi}(\boldsymbol{z} \mid \boldsymbol{X}, \boldsymbol{A})}[\ln p_{\theta}(\boldsymbol{X} \mid \boldsymbol{z}, \boldsymbol{A})] \\ \notag
    &-\mathrm{KL}(q_{\phi}(\boldsymbol{z} \mid \boldsymbol{X}, \boldsymbol{A}) \| p(\boldsymbol{z})).
\end{align}

\begin{figure}[ht]
    \centering
    \includegraphics[width=0.65\linewidth]{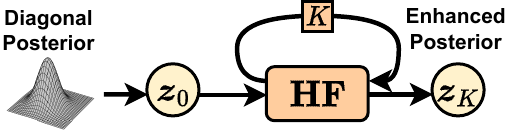} 
    \caption{Illustration of the variational style enhancer.}
    \label{fig:flow}
\end{figure}
To achieve a more flexible posterior distribution other than a simple diagonal Gaussian, we apply the householder-transformation \cite{householderflow_2017} based normalizing flow to enhance the variational inference \cite{vi_2015}. By applying a sequence of invertible mappings $\mathbf{H}^{(k)}, k=1,\ldots, K,$ over the initial variable, we obtain a more valid and flexible probability distribution at the end of this sequence. As shown in Fig. \ref{fig:flow}, $\boldsymbol{z}^{(k)}=\mathbf{H}^{(k)}(\boldsymbol{z}^{(k-1)})$ and the distribution of $\boldsymbol{z}^{(k)}$ can be transformed from the previous $\boldsymbol{z}^{(k-1)}$:
\begin{align}
    p(\boldsymbol{z}^{(k)}) &=p(\boldsymbol{z}^{(k-1)})|\operatorname{det} \frac{\partial {\mathbf{H}^{(k)}}^{-1}}{\partial \boldsymbol{z}^{(k)}}| \\
    &=p(\boldsymbol{z}^{(k-1)})|\operatorname{det} \frac{\partial \mathbf{H}^{(k)}}{\partial \boldsymbol{z}^{(k-1)}}|^{-1},
\end{align}
where $\operatorname{det}$ denotes the Jacobian determinant of the transformation. The density of $\boldsymbol{z}^{(k)}$ is obtained by successively transforming $\boldsymbol{z}^{(0)}$ through a sequence of transformations:
\begin{figure*}[t]
    \centering
    \includegraphics[width=0.8\linewidth]{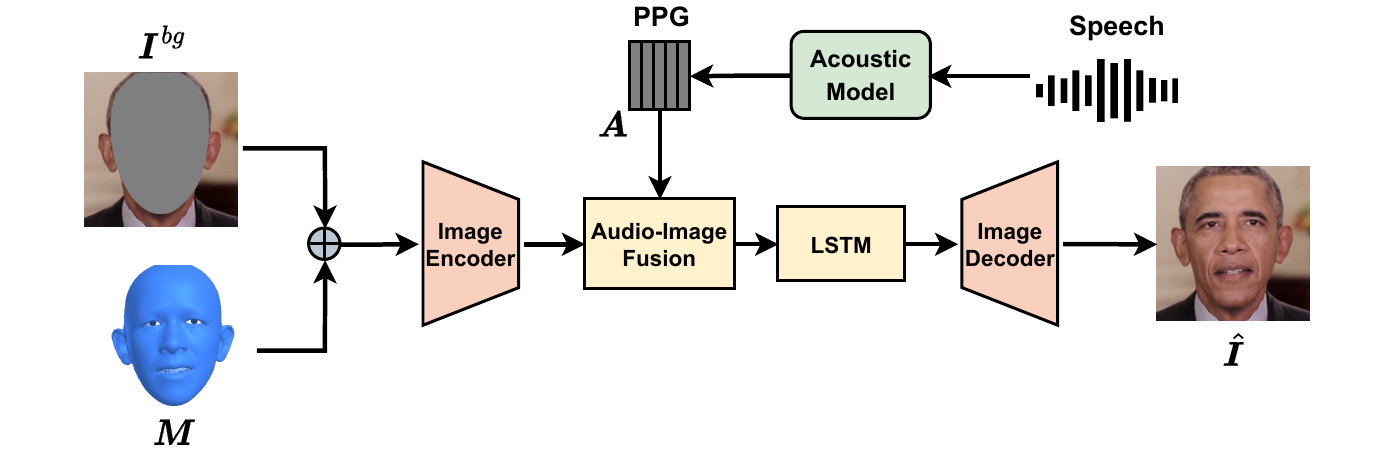} 
    \caption{Illustration of the renderer module. The background images and mesh images are concatenated. The images are fed as a sequence of 16 frames for a training sample. The input can be of variable length in the inference stage.}
    \label{fig:renderer}
\end{figure*}
\begin{align}
    \boldsymbol{z}^{(K)}&=\mathbf{H}^{(K)} \circ \ldots \circ \mathbf{H}^{(2)} \circ \mathbf{H}^{(1)}(\boldsymbol{z}^{(0)}), \\
    \ln p_K(\boldsymbol{z}^{(K)})&=\ln p_0(\boldsymbol{z}^{(0)})-\sum_{k=1}^K \ln |\operatorname{det} \frac{\partial \mathbf{H}^{(k)}}{\partial \boldsymbol{z}^{(k-1)}}|.
    \label{eq:flow}
\end{align}
With the enhanced posterior distribution replacing the vanilla diagonal posterior, Eq. \ref{eq:cvae2_supp} thus becomes 
\begin{align}
    \ln p_{\theta}(\boldsymbol{X} \mid \boldsymbol{A}) &\geq \mathbb{E}_{q_{\phi}({\boldsymbol{z}}^{(0)} \mid \boldsymbol{X}, \boldsymbol{A})}[\ln p_{\theta}(\boldsymbol{X} \mid {\boldsymbol{z}}^{(K)}, \boldsymbol{A}) \\ \notag
    &+\sum_{k=1}^K \ln |\operatorname{det}{\frac{\partial \mathbf{H}^{(k)}}{\partial {\boldsymbol{z}}^{(k-1)}}}|] \\ \notag
    &-\operatorname{KL}(q_{\phi}({\boldsymbol{z}}^{(0)} \mid \boldsymbol{X}, \boldsymbol{A})|| p({\boldsymbol{z}}^{(K)})),
    \label{eq:vae_flow}
\end{align}
where the first reconstruction term is formulated with $\boldsymbol{z}^{(K)}$, since we finally sample from the distribution of $\boldsymbol{z}^{(K)}$. The first term can also be $\mathbb{E}_{q_{\phi}({\boldsymbol{z}}^{(0)} \mid \boldsymbol{X}, \boldsymbol{A})}[\ln p_{\theta}(\boldsymbol{X} \mid {\boldsymbol{z}}^{(0)}, \boldsymbol{A})]$.

\section{Renderer Structure}
We adopt the conditional generative network (CGAN) \cite{cgan_2017} as the basic framework for the renderer.
The eroded background images $\boldsymbol{I}^{bg}$ and corresponding 3D face representation mesh sequence \cite{moai_2021} $\boldsymbol{M}$ are taken as the conditional input. As shown in Fig. \ref{fig:renderer}, the renderer is designed as an encoder-decoder structure. The image encoder is composed of three convolutional layers with stride 2. The encoded bottleneck features contain compact information about face shape, appearance, and image background. To enable the renderer to be perceptual about the relation between speech features and the mouth-region movements and enhance the synthesis accuracy on the mouth shape, the speech features PPG are injected into the bottleneck features. The PPG features and image encoder features are sent into the audio-image fusion module and output the final bottleneck features. This fusion module is constructed with four convolutional layers. To capture the time dependency among the sequence of audio and image features, a long short-term memory (LSTM) \cite{lstm_1997} module is utilized after the fusion module. Finally, the decoder which is composed of two transposed convolutional layers is employed to output the reconstructed images $\hat{\boldsymbol{I}}$.

\section{MOS Guideline}
In the authenticity evaluation, expressiveness evaluation and ablation study, we extensively conduct the MOS tests to verify the effectiveness of the proposed method.
Three main aspects are taken into account in these tests: speech-lip sync, expressiveness \&
richness, and overall naturalness. Fifteen judgers participate in these tests. We now list the questions they are asked to evaluate these three aspects.

\paragraph{1. Speech-Lip Sync.} 
How much do the lip movements match the audio? Very good (5) for no wrong lip movements and have nothing different from the ground-truth person talking. Very poor (1) for the lip movements are totally unreasonable and cannot read content from the lips at all.

\paragraph{2. Expressiveness \& Richness.} 
How vivid or exaggerated is the avatar presented? Very good (5) for rich and contagious expression on the mouth region. Very poor (1) for unreal and monotonous facial movements. 

\paragraph{3. Overall Naturalness.} How real and natural is the synthesized video? Very good (5) for high-quality images and natural avatar appearance. Very poor (1) for fake artifacts or blurred images that can be easily observed.

\end{document}